\documentclass[letterpaper, 10 pt, conference]{ieeeconf}  

\usepackage{xcolor}
\usepackage{graphicx}
\usepackage{amsfonts} 
\usepackage{fancyhdr}
\usepackage{diagbox}
\usepackage[ruled,vlined]{algorithm2e}
\usepackage{amsmath} 
\usepackage{booktabs}
\makeatletter
\usepackage{etoolbox}
\IEEEoverridecommandlockouts                              
\begin{document}
\title{\LARGE \bf Towards Accurate Vehicle Behaviour Classification With Multi-Relational Graph Convolutional Networks }
\author{ Sravan Mylavarapu\textsuperscript{1}\thanks{$^{1}$Authors are with the Center for Visual Information Technology, KCIS, IIIT Hyderabad}, Mahtab Sandhu\textsuperscript{2}\thanks{$^{2}$Authors are with the Robotics Research Center, KCIS, IIIT Hyderabad}, Priyesh Vijayan\textsuperscript{3}\thanks{$^{3}$ School of Computer Science, McGill University and Mila. Work done when the author was at Robert Bosch Center for Data Science and AI.},\\ K Madhava Krishna\textsuperscript{2} , Balaraman Ravindran\textsuperscript{4}\thanks{$^{4}$ Dept. of CSE and Robert Bosch Center for Data Science and AI, IIT
Madras}, Anoop Namboodiri\textsuperscript{1}}

\maketitle
\thispagestyle{empty}
\pagestyle{empty}

\begin{abstract}

Understanding on-road vehicle behaviour from a temporal sequence of sensor data is gaining in popularity. In this paper, we propose a pipeline for understanding vehicle behaviour from a monocular image sequence or video. A monocular sequence along with scene semantics, optical flow and object labels 
are used to get spatial information about the object (vehicle) of interest and other objects (semantically contiguous set of locations) in the scene. This spatial information is encoded by a Multi-Relational Graph Convolutional Network (MR-GCN), and a temporal sequence of such encodings is fed to a recurrent network to label vehicle behaviours. The proposed framework can classify a variety of vehicle behaviours to high fidelity on datasets that are diverse and include European, Chinese and Indian on-road scenes. The framework 
also provides for seamless transfer of models across datasets without entailing re-annotation, retraining and even fine-tuning. We show comparative performance gain over baseline Spatio-temporal classifiers and detail a variety of ablations to showcase the efficacy of the framework.

\end{abstract}

\section{INTRODUCTION}

Dynamic scene understanding in terms of vehicle behaviours is laden with many applications from autonomous driving where decision making is an important aspect to traffic violation. For example, a vehicle cutting into our lane would require us to decide in moving opposite to it (refer to video for applications). 
In this effort, we categorize on-road vehicle behaviours into one of the following categories: parked, moving away, moving towards us, lane change from left, lane change from right, overtaking. To realize this, we decompose a traffic scene video into its spatial and temporal parts. The spatial part makes use of per-frame semantics, object labels to locate objects in the 3D world as well as model spatial relations between objects. The temporal part makes use of flow to track the progress of per frame inter-object relations over time. 

\begin{figure}[t]
    \centering
    \includegraphics[ 
    width=0.89\linewidth,height=0.85\linewidth]{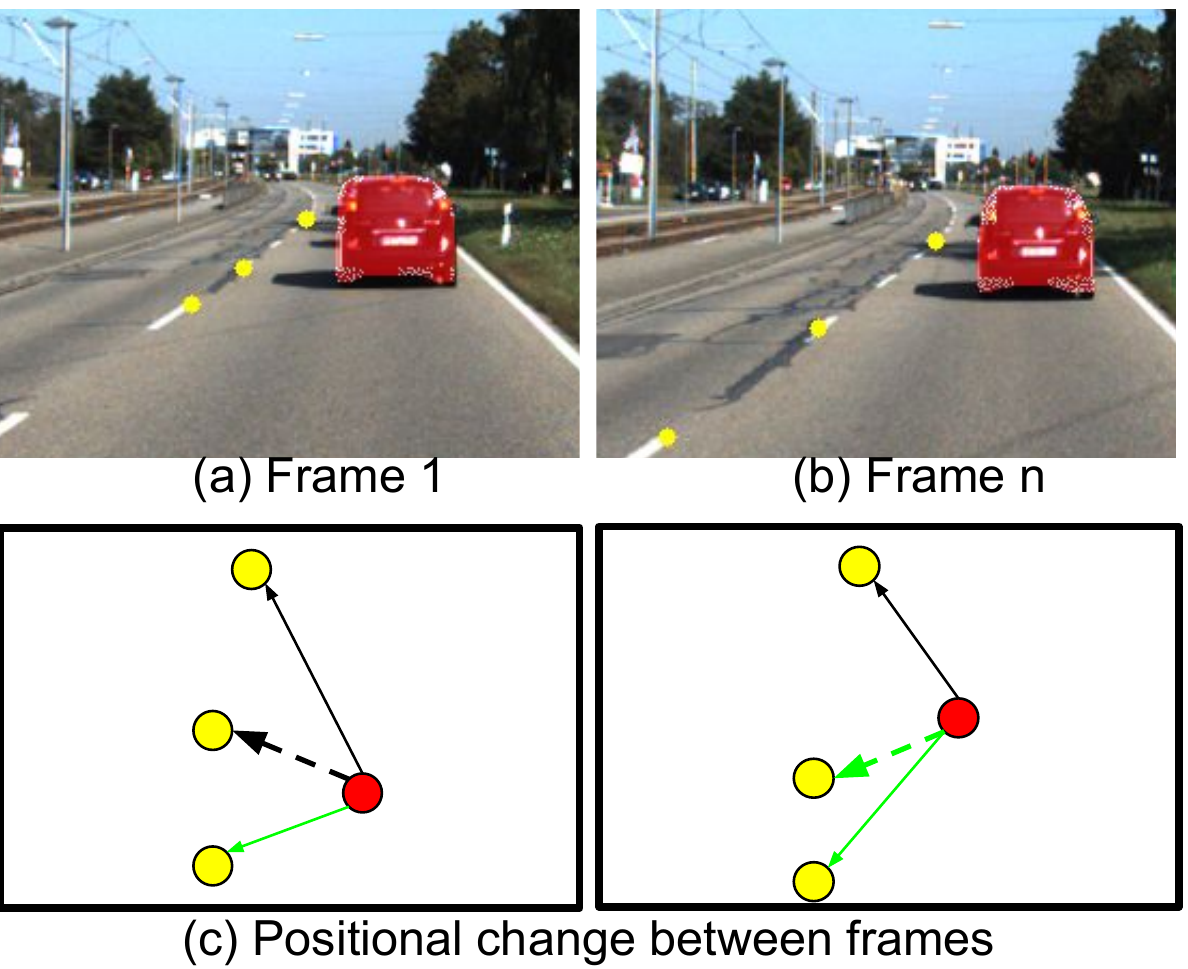}
        
    \caption{\scriptsize The figure depicts how objects (vehicle) behaviour can be modelled as a Spatio-temporal scene graph. The evolution pattern of this scene graph helps us to classify the behaviour. Here the car (red) moves ahead of a lane marking (yellow) . This spatial relation helps us identify that the car is moving ahead}.
    \vspace{-6mm}
    \label{fig:Teaser}
\end{figure}

We use Multi-Relational Graph Convolution Networks (MR-GCN)\cite{schlichtkrull2018modeling} that are now gaining traction to learn per frame embeddings of an object's (object of interest) relation with its surrounding objects in the scene. A temporal sequence of such embeddings is used by a recurrent network (LSTM)\cite{gers1999learning} and attention model to classify the behaviour of the object of interest in the scene. Apart from its high fidelity and seamless model transfer capabilities, the attractiveness of the framework stems from its ability to work directly on monocular stream data bypassing the need for explicit depth inputs such as from LIDAR or stereo modalities. The end to end nature of the framework limits explicit dependencies to the entailment of a well-calibrated monocular camera.
Primarily the proposed framework uses inter-object spatial relations and their evolution over time to classify vehicle behaviours. Intuitively a car's temporal evolution of its spatial relations with its adjacent objects (also called landmarks in this paper) help us, humans, to identify its behaviour.
As seen in Fig. \ref{fig:Teaser} the change of car's spatial relation with lane marking tells us that the car is moving away from us. In a similar vein, an object's relational change with respect to lane markings also decides its lane change behaviour while its changing relations with another moving car in the scene governs its overtaking behaviour. 
The network's ability to capture such intuition is a cornerstone of this effort.

\begin{figure*}
    \centering
    \includegraphics[
    width=0.9\linewidth]{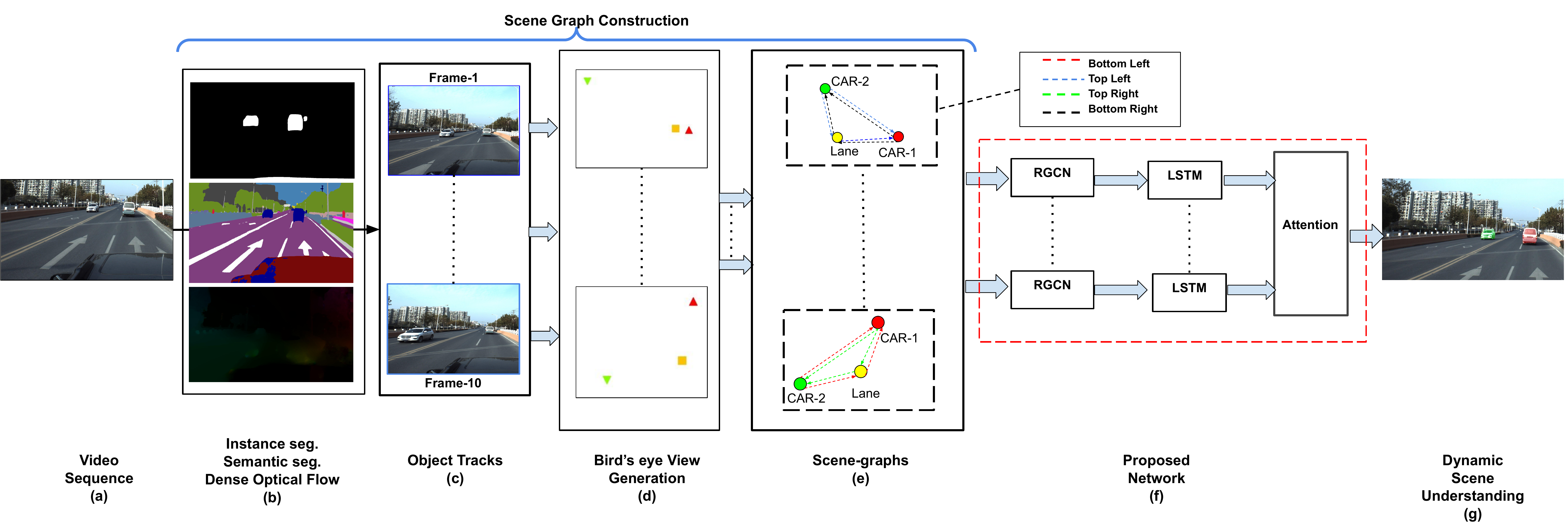}
    \caption{\scriptsize This illustrates the overall pipeline of our framework. The input (a) of the system consists of consecutive frames of a monocular video from cameras mounted on the cars. We use traditional object tracking pipelines, as shown in (b) to extract features.
    These extracted features/tracklets are then transferred into the bird's-eye view as in (d).
    Using these pre-processed features in bird's-eye view, we generate spatial scene graphs as illustrated with examples in (e).
    MR-GCNs are used to encode spatial-relations between objects from this scene graph. Subsequently, this spatial information from MR-GCNs over different frames are passed to an LSTM to model the temporal evolution of such spatial-relations and predict the different vehicle behaviours.
    }
    \label{fig:pipeline}
\end{figure*}

Specifically the paper contributes as follows
\begin{itemize}

    \item We showcase Multi-relation GCN (MR-GCN) along with a recurrent network as an effective framework for modelling Spatio-temporal relations in on-road scenes. Though Graph Convolutional Networks (GCNs) have gained traction recently in the context of modelling relational aspects of a scene, to the best of our knowledge this is perhaps the first such reporting of using an MR-GCN in an end-to-end framework for on-road temporal-scene modelling, ref \ref{sec:GCN}.
    
    \item We reproduce the performance benefits of our framework over a variety of datasets, namely KITTI, Apollo-scape and Indian datasets, to showcase its versatility. Also, our model can effectively transfer its learning seamlessly across different datasets preventing any entailment for annotation, retraining and fine-tuning (as shown in tables \ref{table:main}, \ref{table:transfer})
    
    \item Further results on diverse datasets and various on-road participants (not restricted to cars) such as in Indian roads is another feature of this effort. We also provide ablation studies that show that the model's performance improves with an increase in the number of objects (landmarks) in the scene. Albeit, the model is yet fairly robust and superior to the baseline when only a minimal fraction of the landmarks are available.  
\end{itemize}
\section{RELATED WORK}

\noindent\textbf{Vehicle Behaviour Understanding :}
The closest works to our task are \cite{DBLP:journals/corr/abs-1803-07549, chen2016atomic, sivaraman2013looking}. In \cite{DBLP:journals/corr/abs-1803-07549,sivaraman2013looking} vehicle's behaviour is classified  using a probabilistic model. The classification is with respect to the ego-vehicle network, for example, a car is only to be classified as overtaking if it overtakes the ego car. Whereas in our proposed method, we classify vehicle behaviours with respect to a global frame, i.e. we are able to identify and classify overtaking and other behavioural interactions among any of the objects present in the visible scene.
Methods \cite{schulz2018interaction,sabzevari2016multi}  predict trajectories based on the object's previous interactions and trajectory. These methods are concerned with predicting future motion rather than classifying maneuvers. 
\cite{geng2017scenario} follows a hard rule-based approach for the task of maneuver prediction in contrast to our learnable approach, which is more generic.   
In \cite{chen2016atomic} traffic scenes are  summarized  into textual output. This is in contrast to the proposed framework, which tries to classify objects of interest in terms of a specific behaviour exhibited over the last few evolving time instants.
To our best knowledge the works \cite{jain2015car},  \cite{jain2016structural}, \cite{narayanan2019dynamic}  are the closest to our proposed method, but uses various external sensors such as GPS, face camera and Vehicle dynamics and the classification is for the ego-vehicle alone. In the Computer Vision community, event recognition for stationary surveillance video is well researched over standard datasets \cite{WangCVPR}, \cite{Virat}. Closer home in the robotics community, motion detection methods such as \cite{namdev2012motion,vidal2003optimal,MotDet_Abhijit_IROS2009,fanani2018cnn} some that have included semantics along with motion \cite{reddy2015dynamic, Valada_2017_IROS,chen2016object} bear some resemblance.
However, the proposed method goes beyond the traditional task of motion segmentation. While understanding the presence of motion in the scene, it presents a scalable architecture that recognizes much more complex and diverse set of behaviours by just using monocular videos.\\
\\
\noindent\textbf{Graph Neural Networks :}
Scene understanding has been traditionally formulated as a graph problem \cite{fischler1973representation,marr1978representation}. Graphs have been extensively used for scene understanding, and most of them were solved using belief propagation \cite{felzenszwalb2005pictorial},  \cite{sigal2012loose}. While the success of deep learning approaches into the structured data prediction problems of scene understanding like semantic segmentation and motion segmentation have been successful, there has been relatively sparse literature on scene understanding in the non-euclidean domains like video and behaviour understanding. Recently there have been multiple successful attempts at modelling learning algorithms for graph data and specifically, \cite{NIPS2015_5954,kipf2016semi,DBLP:journals/corr/BrunaZSL13,DBLP:journals/corr/HenaffBL15,DBLP:journals/corr/DefferrardBV16} have proposed algorithms to generalize deep learning to unstructured data. With the success of these methods on the graph classification task, multiple recent works have extended the methods to address multiple 3D problems like Shape segmentation \cite{yi2017syncspeccnn}, 3D correspondence \cite{litany2017deep}, CNN on surfaces \cite{maron2017convolutional}.
We model the relations in a dynamic scene using graph networks to predict the behaviour of the vehicles. The end to end framework wherein the vehicle maps a temporal sequence of spatial relations to scalable number of behaviours is a distinguishing novelty of this effort.

\section{PROPOSED METHOD}
We propose a pipeline to predict maneuver behaviours of different vehicles in a video by modeling the evolution of their Spatio-temporal relations with other objects. The behaviour prediction pipeline is a two-stage process; the first stage involves modeling a video as a series of spatial scene graphs, and the second stage involves modeling this Spatio-temporal information to predict maneuver behaviors of vehicles.

\subsection{Scene Graph Construction}
The scene graph construction is itself a multi-step process where we first identify and track different objects from $T$ consecutive video frames. Next, we orient these objects in a bird's eye view to facilitate the identification of relative positions of various identified objects. Then, for every frame, we encode the relative positional information of different objects with a spatial graph. The steps are elaborated below:

\subsubsection{Object Tracking} \label{sec:preprocess}
For a complete dynamic scene understanding, we use three main feature extraction pipelines as input to identify the different objects and determine the spatial relationships in a video. The major components required for dynamic feature extraction are instance segmentation, semantic segmentation and per-pixel optical flow in a video frame. We follow the method of \cite{MaskRCNN} to compute the instance segmentation of each object. Specifically, the instance segmentation helps detect moving objects and for long term tracking of the object. We follow the method of \cite{rota2018place} to compute the semantic understating of the scene specifically for static objects like poles and lane markings, which act as landmarks. The optical flow is computed between two consecutive frames in the video using \cite{pathakCVPR17learning}. Each of the objects in the scene is tracked by combining optical flow information with features from the instance and semantic segmentation. 

\subsubsection{Monocular to Bird's-eye view}\label{sec:birdseye}
Once we obtain stable tracks for objects in the image space, we move those tracks from image space to a bird's-eye view to better capture the relative positioning of different objects. For the conversion from image to bird's-eye view, we employ Eqn: \ref{egn:birds} as described in \cite{DBLP:conf/cvpr/SongC14} for object localization in bird's-eye view. All the objects are assigned a reference point, for lanes it is the center point of lane marking, for vehicles and other landmarks it is the point adjacent to the road that is assigned as a reference point as seen in Fig. \ref{fig:pipeline} (d). 
Let $b = (x,y,1)$ be the reference point in homogeneous coordinates of the image.
We transfer these homogeneous co-ordinates $b$ to a Bird's-eye view as follows.
\begin{equation}
\label{egn:birds}
\small
    B = (B_x, B_y, B_z) = \frac{-hK^{-1}b}{\eta^{T}K^{-1}b}
\end{equation}
Here $K$ is the camera intrinsic calibration matrix, $\eta$ is the surface normal, and $h$ is the camera height from the ground plane and 
$B$ is the 3D co-ordinate (in the bird's-eye view) of the point $b$ in the  image.

\subsubsection{Spatial Scene Graph }
\label{sec:prop-sgc-graph}
The spatial information in a video frame at time $t$ is captured as a scene graph $S^t$ based on the relative position of different objects in the bird's-eye view. The spatial relation, $S^t_{i,j}$ between a subject, $i$ and an object, $j$ at time, $t$ is the quadrant in which the object lies with respect to the subject, i.e $S^t_{i,j} \in \mathcal{R}_s$, where $\mathcal{R}_s = \{$top left, top right, bottom left and bottom right$\}$. An example of a spatial scene graph is illustrated in Fig. \ref{fig:pipeline} (e). 

\subsection{Vehicle Maneuver Behaviour Prediction}\label{sec:prop-vmb}

We obtain Spatio-temporal representations of vehicles by modeling the temporal dynamics of their spatial relations with other vehicles and landmarks in the scenes over time. The Spatio-temporal representation is obtained by stacking three different neural network layers, each with its own purpose. First, we use a Multi-Relational Graph Convolutional Network (MR-GCN) layer \cite{schlichtkrull2018modeling} to encode the spatial information of each object (node in graph) with respect to its surroundings at time $t$ into an embedding $E^t$. Then, for all nodes, it's spatial embedding from all timesteps, $\{E^1,E^2,..., E^T \}$ is fed into a Long Short-term Memory layer (LSTM) \cite{gers1999learning} to encode the temporal evolution of its spatial information. The Spatio-temporal embeddings, $\{C^1,C^2,..., C^T \}$ obtained at each timestep from the LSTM is then passed through a (multi-head) self-attention layer to select relevant Spatio-temporal information necessary to make behavior predictions. The details of the different stages in the pipeline follow in order below. The overall pipeline is best illustrated in Fig. \ref{fig:pipeline}.

\subsubsection{MR-GCN}
\label{sec:relations}
\label{sec:GCN}
The multi-relational interactions ($S^t_{i,j} \in \mathcal{R}_s$)  between objects are encoded using a Multi-Relational Graph Convolutional Network, MR-GCN \cite{schlichtkrull2018modeling}, originally proposed for knowledge graphs. 

For ease of convenience, we reformulate the representation of each spatial scene graph as $G$, where $G$ is defined as a set of $|\mathcal{R}_s|$ Adjacency matrices, $G=\{A_1,..,A_{|\mathcal{R}_s|}\}$ corresponding to the different relations. Herein, $A_r \in \mathbb{R}^{n\times n}$ with $A_r[i,j]=1$ if $S_{i,j}= r$
otherwise $A_r[i,j]=0, \forall r \in \mathcal{R}_s$; $n$ denotes number of nodes in the graph.

The output of the $k^{th}$ MR-GCN layer, $h^k_G$ is defined below in Eqn: \ref{eqn:MR-GCN}. The MR-GCN layer convolves over neighbors from different relations separately and aggregates them by a simple summation followed by the addition of the node information.
\begin{align}
\small
    \label{eqn:MR-GCN}
    h^k_G = ReLU \big( \sum_{r \in \mathcal{R}} \hat{A}_r h^{k-1}_G W_r^k + h^{k-1}_GW_s^k \big)
\end{align}
where $\hat{A}_r$ is the degree normalized adjacency matrix of relation, $r$, i.e $\hat{A}_r = D_r^{-1}A_r$ where $D_r$ is the degree matrix; $W_r^k$ is the weights associated with the $r^{th}$ relation of the $k^{th}$ MR-GCN layer and $W_s^k$ is the weight associated with computing the node information (self-loop) at layer $k$; For each object, the input to the first MR-GCN layer, are learned object embeddings $\mathcal{E}_o \in \mathbb{R}^{|O|*d}$ corresponding to their object type; $O = \{vehicle, landmark\}$ in our case. Thus, the input layer $h_G^0\in \mathbb{R}^{n*d}$. If $d$ is the dimensions of a $k^{th}$ MR-GCN layer, then $W_s^k,\,W_r^k \in \mathbb{R}^{d*d}$ and $h_G^{k+1} \in \mathbb{R}^{n*d}$.

The MR-GCN layer defined in Eqn: \ref{eqn:MR-GCN} provides a relational representation for each node in terms of its immediate neighbors. Multi-hop representations for nodes can be obtained by stacking multiple MR-GCN layers, i.e., stacking $K$ layers of MR-GCN provides a $K$-hop relational representation, $h^K_G$. In essence, MR-GCN processes the spatial information from each timestep $t$  ($S^t$) and outputs a $K$-hop spatial embedding, $E^t$ with $E^t=h^K_G$.

\subsubsection{LSTM}
To capture the temporal dynamics of spatial relations between objects, we use the popular LSTM to process spatial embeddings of all the objects in the video over time to obtain Spatio-temporal embeddings, $C$. At each time-step, $t$, the LSTM takes in the current spatial embedding of the objects, $E^t$ and outputs a Spatio-temporal embedding of the objects, $C^t$ conditioned on the states describing the Spatio-temporal evolution information obtained until the previous time-step as defined in Eqn: \ref{eqn:LSTM}.
\begin{align}
\small
        \label{eqn:LSTM}
        C^t &= LSTM( E^t, C^{t-1})
\end{align}
\subsubsection{Attention}\label{sec:attention}
We further improve the temporal encoding of each node's spatial evolution by learning to focus on those temporal aspects that matter to understand vehicle maneuvers. Specifically, here we use Multi-Head Self-Attention described in \cite{vaswani2017attention} on the LSTM embeddings. 
The Self-Attention layer originally proposed in the NLP literature, obtains a contextual representation of a data at a time, $t$ in terms of data from other timesteps. The current data is referred to as the query, $Q$, and the data at other time-steps is called value, $V$, and is associated with a key, $K$. Attention function outputs a weighted sum of value based on the context as defined in Eqn: \ref{eqn:att}. 
\begin{align}
\small
        \label{eqn:att}
        Attention_m(\,Q,\,K,\,V\,) = softmax(\dfrac{Q\,K^T}{\sqrt{d_k}})V 
\end{align}

Herein our task, at every time-step, the query is based on the current LSTM embedding, and the key and value are representations based on the LSTM embeddings from other time-steps. For a particular time-step $t$, the attention function looks out for temporal correlations between the Spatio-temporal LSTM embeddings at the current time-step and the rest. High scores are given to those time-steps with relevant information. Herein, we use the multi-head attention, which aggregates (CONCAT) information from multiple attention functions, each of which may potentially capture different contextualized temporal correlations relevant for the end task. A multi-head attention with $M$ heads is defined below:
\begin{align}
\small
        \label{eqn:multi-att}
         &Z^t = \mathop{\parallel}\limits_{m=1}^{m=M} Attention_m(C^t\,W_m^Q , C^t\,W_m^K , C^t\,W_m^V)
\end{align}
where $\parallel$ represents concatenation, $W_m^Q \in \mathbb{R}^{d \times d_k}$  , $W_m^K \in \mathbb{R}^{d \times d_k}$ and $W_m^V \in \mathbb{R}^{d \times d_v}$ are parameter projection matrices for $m^{th}$ attention head. $Z^t \in \mathbb{R}^{n \times Md_v}$ is the concatenated attention vectors for $t^{th}$ time-step from $M$ heads . 

Information from all the time-steps, $Z$, are average pooled along the time-axis and label predictions are made with them. All the three components in the behavior prediction stage, are trained end-end by minimizing a cross-entropy loss defined over the true labels, $Y$ and the predicted labels, $\hat{Y}$. The final prediction components are described below.
\begin{align}
\small
\begin{split}
        \label{eqn:loss}
        &U = Pool_{AVG}(Z) \\
        &\hat{Y} = UW_l \\
        &min ~~ CrossEntropyLoss\,(Y, \hat{Y})
\end{split}
\end{align}
where, $W_l$ projects $U$ to number of classes.  
\begin{figure*}
    \centering
    \includegraphics[trim=0 0 0 20,clip,  width=0.94\linewidth]{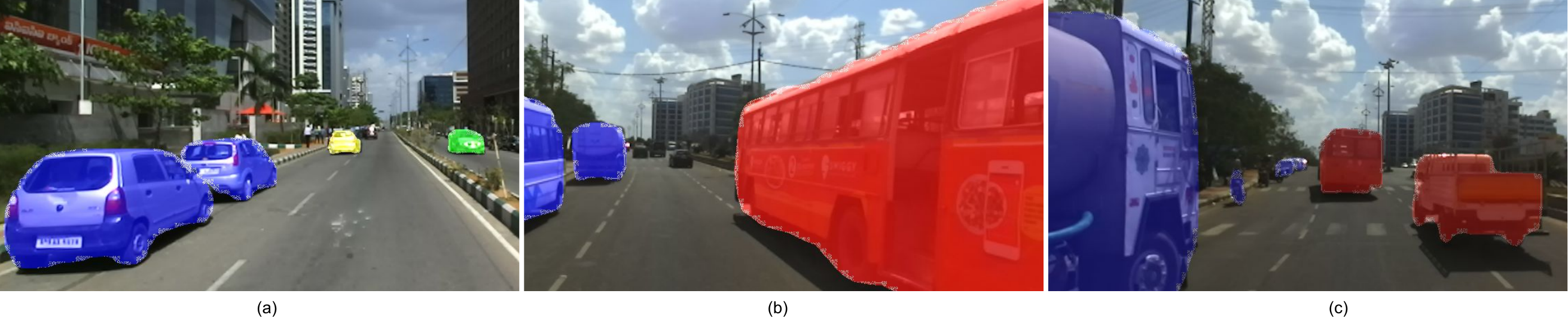}
    \caption{\scriptsize Figure depicts results on the Indian dataset. We can see Vehicle class agnosticism of our Pipeline as non standard Vehicles such as bus and oil-tanker are correctly classified}
    \vspace{-3mm}
    \label{fig:indian-eval}
\end{figure*}

\begin{figure*}
    \centering
    \includegraphics[trim=0 24 0 10,clip, width=0.95\linewidth
    ]{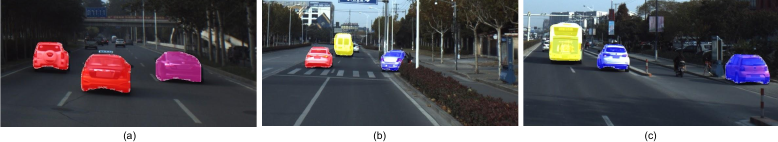}
    \caption{\scriptsize Figures depicts complex Vehicles maneuvers such as overtaking which can be seen in (a), the car on the right represented in magenta color and  lane change which can be seen in image (b) and (c) with yellow on the Apollo dataset}
    \label{fig:apollo_eval}
     \vspace{-3mm}
\end{figure*}

\section{Experiments and Analysis}
We evaluate our vehicle behaviour understating framework on multiple publicly available datasets captured from moving cameras. Code along with hyper-parameters and additional implementation details are available on the project web-page [https://ma8sa.github.io/temporal-MR-GCN/]
\subsection{Training}
All the graphs are constructed for a length of 10 time-steps on all datasets. We empirically found that using 2 layers for MR-GCN provide us with the best results, with first and second layers being 128, 32 respectively. Other variations can be found on our project page. 
The output dimension for Attention and LSTM are equal to their corresponding input.
Training is done on single Nvidia GeForce GTX 1080 Ti GPU.
\subsection{Datasets}
Our Method takes in monocular videos to understand the vehicle behaviour.
We evaluate our framework on three datasets, two publicly available datasets namely, KITTI \cite{Geiger2013IJRR} and Apollo-scapes \cite{huang2018apolloscape} and our proprietary  dataset, Indian dataset which is captured in a more challenging cluttered environment. For all the three datasets we collected  human annotated labels for the vehicle maneuvers.

\subsubsection{Apollo-scape Dataset}
 Apollo-scape harbors a wide range of driving scenarios/maneuvers such as \textit{overtaking}, \textit{Lane change}. We selected this dataset as our primary dataset because it contains  considerable instances of complex behaviours such as \textit{overtaking}, \textit{Lane change} which are negligible in other datasets.

\subsubsection{KITTI Dataset}
This dataset is constrained to a city and regions around the city along with few highways. We pick sequences 4, 5, 10 for our purpose.


\subsubsection{Indian Dataset}
To account and infer the behaviour of the objects with high variation in object types, we capture a dataset in the Indian driving conditions. This dataset contains Non-Standard vehicles which are not previously modeled by the framework. 
\vspace{-1mm}
\subsection{Qualitative Analysis}
\begin{figure*}
    \centering
    \includegraphics[
    clip ,width=0.9\linewidth]{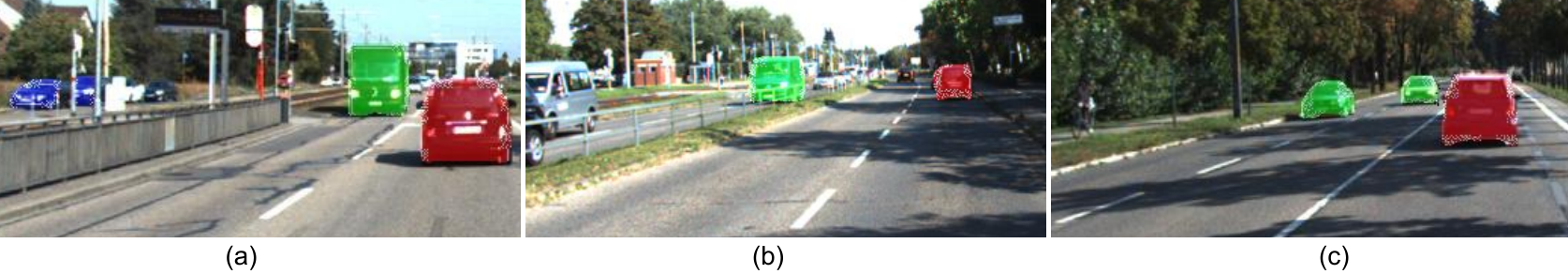}
    \caption{Figure depicts Results obtained on KITTI Tracking dataset.}
    \vspace{-5mm}
    \label{fig:kitti-eval}
\end{figure*}

The following color coding is employed across all qualitative results. Blue denotes vehicles that are \textit{parked}, red is indicative of vehicles that are \textit{moving away}, and green the vehicles that are \textit{moving towards us}. Yellow indicates \textit{lane changing: left to right}, orange indicates \textit{lane changing: right to left} behaviour with magenta denoting \textit{overtaking}. 
Complex behaviours such as overtaking and changing lanes can be observed in Fig. \ref{fig:apollo_eval} which shows qualitative results obtained on Apollo dataset. Fig. \ref{fig:apollo_eval} (a) depicts a car (right most car) in the midst of overtaking (seen with magenta color). Fig. \ref{fig:apollo_eval} (b) and \ref{fig:apollo_eval} (c) present us with cases of lane change. In Fig. \ref{fig:apollo_eval} (b) we see a van changing lane, along with a moving car, In Fig. \ref{fig:apollo_eval} (c) we observe a bus changing lane from left to right along with two cars parked on the right side. Object class agnosticism can be observed in Fig. \ref{fig:indian-eval} (Indian dataset), where behaviour of several non-standard vehicles are accurately predicted. In Fig. \ref{fig:indian-eval} (a) we can see a car on the  left side of the road (predicted \textit{changing lane}) coming onto the road along with parked cars and one car coming towards us. In Fig. \ref{fig:indian-eval} (b) truck and bus are \textit{parked} on left side of the road while another bus is accurately classified as \textit{moving forward} and not \textit{overtaking}. Note that behaviour is classified as overtaking only if a moving vehicle overtakes (moves ahead of) another moving Vehicle.
Fig. \ref{fig:indian-eval} (c) provides a case with a semi-truck and bus  moving away from us, along with a truck parked. Fig. \ref{fig:kitti-eval} shows qualitative results obtained on KITTI sequences 4, 5, 10.

\subsection{Quantitative Results}
\label{sec:res-quant}
In-depth analysis of our model's performance is available in the form of confusion matrix, ref table \ref{table:confusion}. 
To show that our method is sensor invariant we evaluate our model on KITTI and Indian datasets, results are tabulated in table  \ref{table:transfer}. We provide ablation studies on model as well as land marks and compare our model with  few baselines.

\subsubsection{Ablation study on LandMarks}
\label{sec:res-landmark}
Landmarks are the stationary points (lane marking, poles, etc.) which along with other vehicles are used by MR-GCN to  encode a spatially-aware embedding for a node. We carry out an ablation study to understand the effects of how the number of landmarks available in a sequence affects the performance of our Pipeline. We train three models, one with 50 percent landmarks available, another with 75 percent available and the last one with all the landmarks available. The results mentioned in above experiments are tabulated in table \ref{table:main} under their respective rows  \textit{ours(50\%)},  \textit{ours(75\%)}, \textit{ours(full)}. Behaviours which are directly dependent on landmarks (lane change, moving ahead and back) show a difference of atleast 10\% when comparing the full model and the model trained with 75\% landmarks and a difference of atleast 15\% when compared with a model trained with 50\% landmarks. The difference is less significant for overtaking as its influenced by other moving vehicles. We observed that with less number of landmarks, the network is biased towards 'parked' label, hence the accuracy of it remains high, while others reduce. The observed results make it evident that the proposed model is highly robust even with fewer landmarks and also, that performance improvement is achievable with the utilization of more information in the form of increased landmarks.
\subsubsection{Ablation study on Model}
\label{sec:res-Model_ablation}
To study the importance of each component of our system, we perform an ablation study on the model.
We observe the importance of encoding spatial information along with temporal aspects by comparing variations of our network with each other, as described in
table \ref{table:ablation}. First, we use MR-GCN followed by a simple LSTM (G + L) which provides below par results but performs comparatively better than MR-GCN followed by a single-head attention (G + SA). Adding a single head Attention to MR-GCN and LSTM (G + L + SA), helps to weight the spatio-temporal encodings as described in
\ref{sec:attention} providing a huge improvement in accuracies. The final model with multi head attention (G+L+MA) attains improvement specially for $overtake$ showing it's complex behaviour. 
Overall, each component in the network provides a meaningful representation based on the information encoded in it. 

\noindent \textbf{Baseline comparison : }
\label{sec:res-SVM}
We observe the importance of spatial information encoded as a graph by comparing the performance of our method (G+L+MA)  against a simple LSTM (L) and a LSTM followed by Attention (L + MA) models,
which encode relational information with a 3D location based positional features; ref table \ref{table:ablation}.
To train these baseline models (L and L + MA), we give 3D locations of objects obtained from the bird's eye view (\ref{sec:birdseye}) as a direct input. For each object in the video, we create a feature vector consisting of its distance and angle with all other nodes for $T$ time-steps. The distance is a simple \textit{Euclidean distance}. To account for the feature of an object (lane-markings/Vehicles), we create a 2 dimensional one-hot vector \{1,0\} representing Vehicles and \{0,1\} representing static objects. To create a feature vector for the $i^{th}$ object, we find distances and angles with all other nodes for $T$ time-steps in the scene. In both the cases, the above features are fed to the LSTM at each time-step. We pool the output from LSTM for L and output from Attention layer for L + MA, along time dimension and project to number of classes by using a dense layer. \\
Table \ref{table:ablation} shows clear improvement in terms of accuracies for models having a MR-GCN to encode the spatial information when compared to the simple recurrent models (L and L + MA). Our method outperforms the baselines by a good margin for all classes.

The complete method and algorithm for creating feature vectors and training the models are described in detail in the project web page. [https://ma8sa.github.io/temporal-MR-GCN/]. 

\begin{table}[t]
\renewcommand{\arraystretch}{1.3}
    \centering
    \begin{tabular}{|c|c|c|c|c|c|c|c|}
    \hline
    \diagbox[width=2cm]{GT}{Predicted } & MVA & MTU & PRK & LCL & LCR & OVT & total \\
    \hline
    MVA  & \textbf{691} & 1 & 13 & 16 & 4 & 83 & \textbf{808} \\ 
    \hline
    MTU  & 1 & \textbf{212} & 21 & 0 & 2 & 0 & \textbf{ 236} \\ 
    \hline
    PRK  & 8 & 39 & \textbf{1330} & 19 & 7 & 10 & \textbf{1413} \\ 
    \hline
    LCL & 10 & 0 & 6 & \textbf{137} & 0 & 8 & \textbf{161} \\ 
    \hline
    LCR  & 6 & 0 & 5 & 3 & \textbf{112} & 3 & \textbf{129} \\ 
    \hline
    OVT  & 18 & 0 & 1 & 0 & 0 & \textbf{53} & \textbf{72} \\ 
    \hline
    \hline
    \end{tabular}
    \caption{\scriptsize 
    confusion matrix for model trained on Apollo dataset.
    Due to space constraints we have used abbreviations for label names, MVA : moving away from us, MTU: moving towards us, PRK: parked, LCL: lane change left to right, LCR: lane change right to left, OVT: overtaking and GT : Ground truth}
    \label{table:confusion}
    \vspace{-6mm}
\end{table}


\begin{table}[t]
\renewcommand{\arraystretch}{1.3}
    \centering
    \begin{tabular}{|c|c|c||c|c|}
    \hline
    Trained on & \multicolumn{2}{|c||}{Apollo} &  KITTI &  Indian \\
    \hline
    Tested on & KITTI & Indian & KITTI & Indian \\
    \hline
    Moving away from us   &99 & 99 & 85 & 85\\
    Moving towards us  &  98 & 93 & 86 &  74  \\
    Parked   & 99 & 99 & 89 & 84 \\
    \hline
    \end{tabular}
    \caption{\scriptsize This showcases the transfer learning capabilities of our Method. Values provided are accuracies}
    \vspace{-6mm}
    \label{table:transfer}
\end{table}


\begin{table}[t]
\vspace{2.5mm}
\renewcommand{\arraystretch}{1.3}
    \centering
    \begin{tabular}{|c|c|c|c|}
    \hline
Tested on & \multicolumn{3}{|c|}{Apollo-Scape} \\
    \hline
     Method  & \shortstack{ours \\ (50\%)}& \shortstack{ours \\ (75\%)} & \shortstack{ours \\ (full)} \\
    \hline     
    Moving away & 72 & 76 & \textbf{85.3} \\ 
    Moving towards us & 67 & 75 &  \textbf{89.5} \\ 
    Parked & 97 & 97 & \textbf{94.8} \\
    LC left - right & 66 & 74 & \textbf{84.1} \\
    LC right - left & 71 &72 & \textbf{86.4} \\
    Overtaking & 68 & 65 & \textbf{72.3} \\
    \hline
    \end{tabular}

\caption{\scriptsize Comparison between models trained with 50\% of landmarks and 75\% and 100\% of landmarks. Values provided are accuracies.}

    \label{table:main}
    \vspace{-6mm}
\end{table}

%


\begin{table}[t]
\renewcommand{\arraystretch}{1.3}
    \centering
    \begin{tabular}{|c|c|c|c|c|c|c|}
    \hline
    Classes & MVA & MTU & PRK & LCL & LCR & OVT \\
    \hline
    Architecture &  &  &  &  &  &  \\
    \hline
    G + L + MA & \textbf{85} & \textbf{89 }& \textbf{94} & \textbf{84} & \textbf{86} & \textbf{72} \\ 
    G + L + SA & 84 & 75 & 95 & 51 & 65 & 51 \\ 
    G + L & 78 & 72 & 74 & 37 & 49 & 41 \\
    G + SA & 70 & 18 & 54 & 38 & 14 & 40 \\
    
    \hline
    L & 37 & 35 & 34 & 6 & 24 & 15 \\
    L + MA & 56 & 41 & 46 & 8 & 27 & 13 \\
    \hline
    \end{tabular}

\caption{\scriptsize  Baselines and ablation study on the model trained on Apollo. 
Abbreviations for model architectures (rows) are, G : MR-GCN, L : LSTM, MA : Multi-Head Attention, SA : Single-Head Attention. Nomenclature in the architecture names reflect the components in it. column abbreviations are similar to table \ref{table:confusion}. All the values provided are accuracies.}
    \label{table:ablation}
    \vspace{-10mm}
\end{table}





\subsubsection{Transfer Learning}
\label{sec:res-transfer}
As embeddings obtained from MR-GCN are agnostic to the visual scene, they are dependent on the scene graph obtained from the visual data (images).  
To show this, we trained the model only on Apollo dataset and tested it on Indian and KITTI. While testing, we removed the lane change and overtaking behaviour as it is not present in KITTI and Indian dataset. In table \ref{table:transfer}, we observe that results are better with the model trained on Apollo dataset as compared to models trained on their respective datasets. This could be attributed to the presence of more complex behaviours (overtaking, lane change) in Apollo and shear difference in size of datasets (Apollo has 4K frames as compared to 651 and 722 frames in Indian and KITTI datasets).  
\section{Conclusion}
This paper is the first to pose the problem of classifying a Vehicle-of-interest in a scene in terms of its behaviours such as "Parked", "Lane Changing", "Overtaking" and the like by leveraging static landmarks and  changing relations. The main highlight of the model's architecture is its ability to learn behaviours in an end-to-end framework as it successfully maps a temporal sequence of evolving relations to labels with repeatable accuracy. 
The framework helps in detecting  traffic violations such as overtake on narrow bridges and in deciding to maintain safe distance from a driver (aggressive) changing his state frequently.
The ability to transfer across datasets with high fidelity and robustness in the presence of a reduced number of objects and an improved 
performance over base-line methods summarizes the rest of the paper.
\section{Acknowledgement}
The work described in this paper is supported by  MathWorks. The opinions and views expressed in this publication are from the authors, and not necessarily that of the funding bodies.




\bibliographystyle{IEEEtran}
\bibliography{references2.bib}

\end{document}